\pdfoutput=1

\documentclass[kdmile,a4paper]{kdmile} 
\usepackage{graphicx,url}  
\usepackage[T1]{fontenc}   
\usepackage{subcaption}

\newdef{definition}[theorem]{Definition}
\newdef{remark}[theorem]{Remark}

\title{Classification of User Reports for Detection of Faulty Computer Components using NLP Models: A Case Study\\}

\author{Maria de Lourdes M. Silva, André L. C. Mendonça, Eduardo R. D. Neto, Iago C. Chaves, \\ Felipe T. Brito, Victor A. E. Farias, Javam C. Machado }

\institute{Universidade Federal do Ceará, Brazil \\ 
\email{\{malu.maia, andre.luis, eduardo.rodrigues, iago.chaves, \\ felipe.timbo, victor.farias, javam.machado\}@lsbd.ufc.br}
}

\begin{abstract}
Computer manufacturers typically offer platforms for users to report faults. However, there remains a significant gap in these platforms' ability to effectively utilize textual reports, which impedes users from describing their issues in their own words. In this context, Natural Language Processing (NLP) offers a promising solution, by enabling the analysis of user-generated text. This paper presents an innovative approach that employs NLP models to classify user reports for detecting faulty computer components, such as CPU, memory, motherboard, video card, and more. In this work, we build a dataset of $341$ user reports obtained from many sources. Additionally, through extensive experimental evaluation, our approach achieved an accuracy of 79\% with our dataset.

\end{abstract}



\begin{document}


\maketitle

\section{Introduction}

The advancements in electronic fabrication technologies have led to a large-scale production of computer components. However, it is common for these components to present faults over time. This fact has significantly increased the need for robust diagnostic and performance assessment systems to ensure reliability and efficiency of these devices. 


Computer manufacturers have responded to this demand by providing users with tools to report faults. For example, \textit{Lenovo Diagnostics}\footnote{https://support.lenovo.com/us/en/solutions/diagnose} offers a comprehensive suite of tests to diagnose problems with various components, enhancing the user experience by identifying hardware issues quickly and accurately. Similarly, tools like \textit{Dell SupportAssist}\footnote{https://www.dell.com/pt-br/lp/dt/support-assist-for-pcs} and \textit{HP PC Hardware Diagnostics}\footnote{https://support.hp.com/br-pt/help/hp-pc-hardware-diagnostics} perform detailed system scans, run performance tests, and provide actionable insights to resolve hardware issues promptly, reducing downtime and the need for direct customer support. 

Despite these advancements, there remains a significant gap in the ability of fault-reporting platforms to effectively utilize textual reports provided by users. Incorporating textual reports into such tools is crucial for several reasons. Firstly, users can provide detailed descriptions of the problems they are experiencing, which can help pinpoint the exact faulty component. In addition, textual reports facilitate better communication between the user and the diagnostic system, by allowing users to describe their issues in their own words. 


In this context, Natural Language Processing (NLP) offers a promising solution. It can analyze and interpret textual reports, extracting key information and correlating it with known hardware issues. By leveraging NLP, fault-reporting platforms can better understand the context and specifics of user-reported problems, leading to more accurate fault detection and quicker resolutions. 

This paper leverages NLP models based  on Bidirectional Encoder Representations from Transformers (BERT) \cite{devlin2018bert} and sentence-transformers \cite{reimers2019sentence} to classify user reports for the detection of faulty computer components. For instance, consider a user interacting with a fault-reporting platforms. The user might type, ``\textit{My laptop screen flickers when I adjust the angle}.'' The NLP model analyzes this input and identifies the issue as potentially related to the video card. The system then suggests running a targeted diagnostic module for the video card, providing a simplified and efficient troubleshooting process.

In particular, our contributions are: 

\begin{itemize}
    \item We create a labeled dataset containing user reports related to eight specific types of faulty computer components: video card, storage, motherboard, battery, audio, CPU, memory, and network.
    \item We develop NLP models to classify the collected user reports for detecting faulty computer components by adopting machine learning paradigms such as zero-shot and few-shot learning with transformer-based language models.
   \end{itemize}

The rest of this paper is structured as follows: Section~\ref{sec:related} reviews the main related work in the field. Section~\ref{sec:database} describes the methodology used for constructing the dataset the subsequent NLP model development process. The analysis and evaluation of our experiments are presented in Section~\ref{sec:experiments}. Finally, Section~\ref{sec:conclusion} concludes the paper and outlines future research directions.

\section{Related Work} 
\label{sec:related}


In this section, we survey the literature for related work on text reports for defects/faults on computing systems. The classification of users' reports for defective product identification has been addressed in \cite{abbas2023defective}. This work explores machine learning models with psychological, linguistic and discrete emotions features on a crawled dataset from Amazon with accuracy of $84\%$. The classification of defect types on server development projects were investigated in \cite{su2017creating}. This work classify testers' reports on BIOS defects using decision tree C4.5, Naive bayes, bayes net, logistic regression, neural network with $80\%$ to $82\%$ of accuracy. Also, \cite{sohrawardi2014comparative} tackled the problem of classifying user reports submitted to Bug Tracking Systems with respect to whether or not the report is a bug or not. The text is represented using traditional representation bag-of-words + TF-IDF after stop word removal and stemming. They applied naive bayes, k-nearest neighbors, support vector machine, nearest centroid classifier and multilayer perceptron network for prediction which results in an average error rate of $23.68\%$. 

\cite{alonso2019label} addresses the automatic labeling of unlabeled issues in the issue tracking system of open-source projects on GitHub. The possible labels are: bug, duplicated, enhancement, help wanted, question, invalid and wont-fix. They used the bag-of-word + TF-IDF for represeting issues. As machine learning models, they experimented the support vector machine and the naive Bayes Multinomial achieving a average AUROC of $0.9$. Similarly, transformer-based language models were used to predict the label of user-reported bugs from many github repositories of mobile applications. This approach obtained accuracy varying from $95\%$ to $96.7\%$. Thus, it provides evidence that transformer-based language models can be a promising approach for text classification of user reports. However, to the best of our knowledge, none of the works in the literature explores faults on computer components using transformer-based language models.

Additionally, the evaluation of different language models for specific tasks has been explored in previous studies~\cite{almeida2024evaluation,zhu2023dyval,jahan2024comprehensive}. These works assessed various models in areas such as information extraction from digitized documents, reasoning tasks (including logical, mathematical, and algorithmic problems), and biomedical text processing.

\section{Methodology}
\label{sec:database}

The following sections describe the methodology used to construct the dataset, the criteria for model selection, and the procedures followed in conducting the experiments.

\subsection{Dataset creation}

The ``User Report of Failed Components'' dataset, or simply ``User Report'' dataset, was created for academic research and industrial purposes. It contains two columns: \textit{content} and \textit{label}. The \textit{content} column contains user reports detailing issues encountered while using the devices. The \textit{label} column identifies the device experiencing the problem, i.e., the computer component. This data can be utilized to train text classification models in identifying faulty computer components.

\begin{figure}[!htbp]
    \centering
    \includegraphics[width=\textwidth]{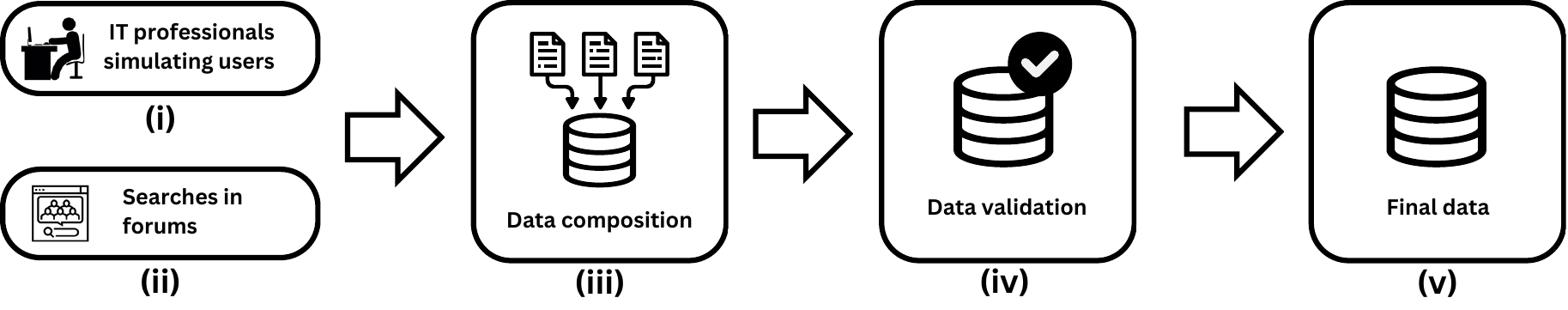}
    \caption{Steps for creating the ``User Report'' dataset.}
    \label{fig:dataset-steps}
\end{figure}

Figure \ref{fig:dataset-steps} describes the steps for creating the ``User Report'' dataset. In the first step (i), we had the participation of 32 IT professionals with extensive experience in developing tools for computer hardware diagnostics. These tools perform various tests to provide feedback to the user about the computer's health status. Each developer had to fill out a form proposing various 
sentences with possible issues for different computer components. The idea is to simulate the user's interaction with IT support when reporting issues on their computers. 


The second step (ii) involved conducting web searches on specialized hardware forums, company FAQs (frequently asked questions), and user guides for hardware components. We filtered the results to identify different devices and the potential issues reported by the online community. This procedure was conducted manually, where we searched for defect reports related to computer components. For instance, we accessed the NVIDIA forum to look for reports of malfunctioning graphics cards.

For steps (i) and (ii), we focused our study on eight groups of computer components: audio, network, motherboard, CPU, memory, battery, storage, and video card. The selected modules cover a substantial portion of the components found in a computer system. 
In the third step (iii), we compiled the collected data in a unique dataset. Finally, in the last step (iv), we performed the validation step, removing data inconsistencies, such as duplicates and the use of a label that does not correspond to any of the eight groups of device modules.

\subsection{Models Selection} 
\label{sub:models-selection}

In this section, we discuss the criteria for selecting the language models for experimentation. We selected models that are enhanced versions of the well-known BERT and that have high accuracy using the Multi-Genre Natural Language Inference dataset MNLI~\cite{williams2017broad} test sets. Table \ref{tab:approaches-info} summarizes the approaches used in this work.



For the zero-shot classification task, we selected two models: BART \cite{lewis2019bartdenoisingsequencetosequencepretraining} and DeBERTa \cite{he2021debertadecodingenhancedbertdisentangled}. The DeBERTa approaches that we selected have four different training configurations, so we consider each configuration as a different approach. 
The BART Large (BL) approach includes 400 million parameters to improve performance in complex tasks compared to traditional BART. It is pre-trained with (MNLI)~\cite{williams2017broad}, a bitext classification task dataset that predicts if one sentence entails another. It concerns classifying pairs of texts, where the goal is to predict if the first sentence implies or entails the second sentence. 

DV3 is an improvement of the traditional model DeBERTa. It incorporates some enhancements to improve the performance of NLP tasks, such as using multiple NLP benchmarks to train the model. Although it has 184M parameters, the model is designed to be parameter-efficient, providing high performance without requiring many parameters. The main difference between DV3 and DV3M lies in their training and fine-tuning. DV3 is designed for English language tasks, while DV3M is a multilingual model, which enables it to capture the context in multiple languages. It is fine-tuned on a cross-lingual dataset, making it suitable for cross-lingual natural language inference tasks. Both approaches share the same architecture and model size.

DV3X is another variant of DeBERTa, and it is very similar to DV3M, but it mainly focuses on cross-lingual natural language Inference (XNLI)~\cite{conneau2018xnli} tasks, providing a superior performance due to the large training dataset. This version is designed to handle multiple languages and is optimized for NLI tasks across those languages. It also shares the same architecture of DV3 and DV3M. 
Similarly to the other DeBERTa approaches, DV3T has 184M parameters. The difference is that DV3T is fine-tuned on specific NLI datasets that compose a tasksource dataset~\cite{sileo2023tasksource} to optimize its performance for particular tasks. The tasksource includes multilingual, cross-lingual, and English data.

\begin{table}[t] \footnotesize
    \centering
    \caption{Configuration of each approach described in this section.}
    \begin{tabular}{cccc} \hline
         \textbf{Acronym} & \textbf{Model} & \textbf{Model size (\# Parameters)} & \textbf{Training data} \\ \hline
         BL & BART & 407M & English data \\
         DV3 & DeBERTa & 184M & English data \\
         DV3M & DeBERTa & 184M & Multilingual\\
         DV3X & DeBERTa & 184M & Cross-lingual \\
         DV3T & DeBERTa & 184M & Tasksource \\ 
         6MLM & MiniLM & 22.7M & English \\
         12MLM & MiniLM & 33.4M & English \\
         MP & MPNet & 109M & English \\
         MMP & MPNet & 278M & Multilingual \\
         MPQA & MPNet & 109M & Multilingual \\ \hline
    \end{tabular}
    \label{tab:approaches-info}
\end{table}

The models selected for few-shot learning are the sentence transformers~\cite{reimers2019sentence}, specifically, MiniLM~\cite{wang2020minilm} and MPNet~\cite{song2020mpnet}. Unlike traditional transformers models, sentence transformers can take an entire sentence or paragraph as input rather than focusing on individual words. It also includes new tasks like large-scale semantic similarity comparison, clustering, and information retrieval. 
Two of our baseline approaches use the MiniLM model. The difference between them is the number of transformers layers. The 6-layer MiniLM (6MLM) is faster in terms of inference, but it might not capture complex patterns as effectively as the 12-layer MiniLM (12MLM), which generally achieves higher accuracy and performs better across a wide range of NLP tasks. 

On the other hand, the approaches that use MPNet model are distinct regarding the training data. MPNet-base (MP) is trained with English data, so it performs well for texts in English and may not be accurate for other languages. The other MPNet-based approach is the Multilingual MPNet (MMP), which supports multiple languages and is highly effective in multilingual semantic similarity and paraphrase identification. Finally, the last MPNet-based approach is the MPQA, trained with a dataset that contains 215 million question-and-answer pairs from different sources.  

\subsection{Retraining Step}
In this section, we describe the NLP models' retraining settings. The user report dataset was partitioned twice to get the training and test subsets used in experimental analysis. The two partition steps are defined below.

\begin{itemize}
    \item [1.] We split the dataset into two sets to stratify the data and get a balanced proportion of samples of each class. Let's call the partitions of the first split P1 and P2. P1 contains $70\%$ of the user report data, and P2 contains $30\%$. The test set is composed of the entire P2.
    \item [2.] We define the data used to retrain the pre-trained models in the second split. This training data is a subset of P1. It has a variable size since it relies on the employed learning technique. For example, the training set contains $k\geq1$ samples of each class in few-shot learning, but the training set has zero samples in zero-shot learning.
\end{itemize}

We retrain the approaches described in Section~\ref{sub:models-selection} using the P1 set. Then, we evaluate the approaches using the test set P2 by comparing their accuracy, F1 score, and the resulting confusion matrix with normalized values. This evaluation process involved sampling different training and test sets, running each approach fifteen times, and averaging the values of each metric computed over the models' predictions to avoid biased results.

The training configuration employed is described as follows. Zero-shot learning (ZSL) approaches used no sample to retrain the models. In one-shot learning (OSL) approaches, we ran fifty epochs with batches containing four samples, and each computer component had one sample addressed for the model's retraining. In few-shot learning approaches, we ran fifty epochs with batches containing twelve samples, and each computer component had $k=23$ samples addressed for retraining. We use AdamW optimizer~\cite{loshchilov2018fixing}, which is a version of Adam with weight decay, with a learning rate of $0.001$, and $\beta_1=0.9$ and $\beta_2=0.999$.

\vspace{-.2cm}

\section{Results}
\label{sec:experiments}

In this section, we empirically evaluate the effectiveness of the proposed approach. The experiments were conducted on the Google Colab platform using a system with 62GB of RAM, 201GB of storage, and the NVIDIA L4 GPU with 22.5GB of RAM.

\vspace{-.2cm}

\subsection{Dataset}
\label{subsec:data-analysis}


A sample of the newly created ``User Report'' dataset is showed in Table \ref{tab:reports}. For instance, in the first line of Table \ref{tab:reports_values}, we have the sentence ``\textit{I have a power issue.}'' as the content, which is an issue regarding the Battery module device. In total, we collected 341 reports, with 51 records regarding the video card module, 48 to the storage module, 47 to the motherboard, 43 to the battery, 41 to the audio, 39 to the CPU, 39 to the memory, and finally, 33 to the network module, as summarized in Table \ref{tab:reports_values}.


\subsection{Models Evaluation}
\label{sec:models_evaluation}

\begin{figure}
\vspace*{-3mm}
    \caption{Confusion matrices of each approach's prediction.}
    \label{fig:confusion-matrices}
    \begin{minipage}[t]{.32\textwidth}
        \centering
        \includegraphics[width=1.\textwidth]{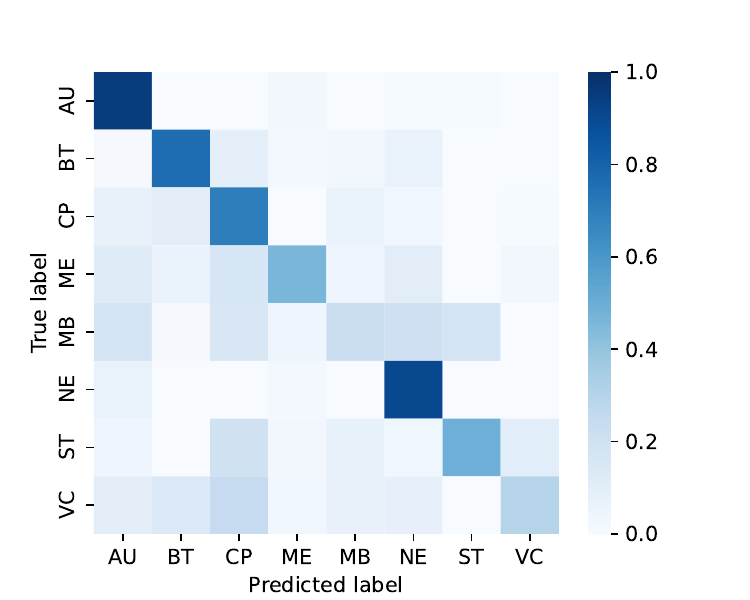}
        \subcaption{ZSL BL}\label{fig:cm-bart-large}
    \end{minipage}
    \begin{minipage}[t]{.32\textwidth}
        \centering
        \includegraphics[width=1.\textwidth]{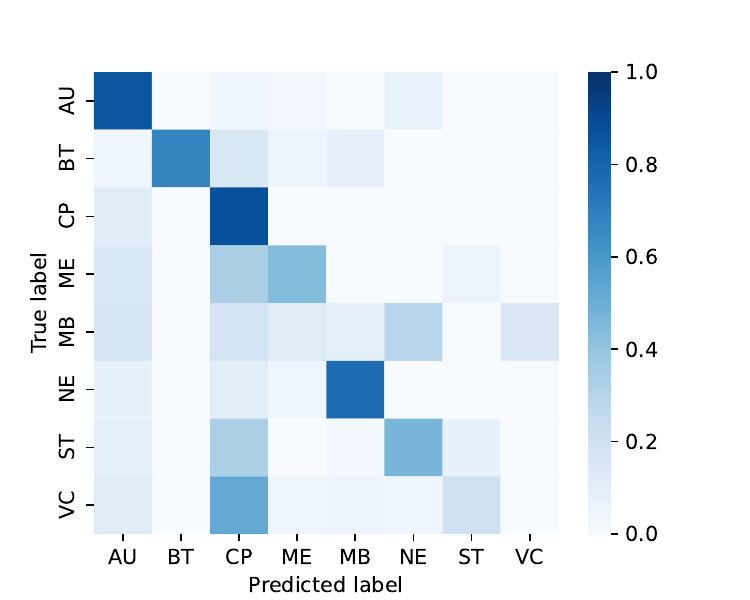}
        \subcaption{ZSL DV3}\label{fig:cm-dv3}
    \end{minipage}
    \begin{minipage}[t]{.32\textwidth}
        \centering
        \includegraphics[width=1.\textwidth]{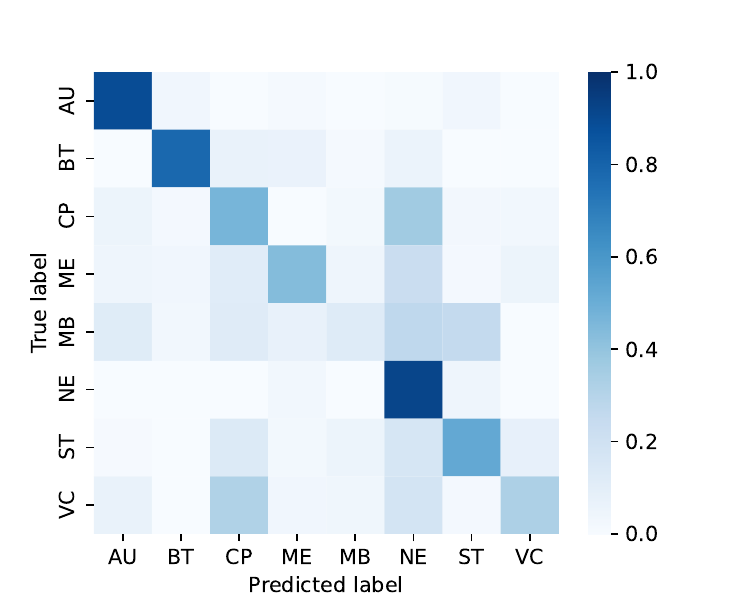}
        \subcaption{ZSL DV3M}\label{fig:cm-dv3m}
    \end{minipage}
    \hfill
    \vspace*{-3mm}
    \center
    \begin{minipage}[t]{.32\textwidth}
        \centering
        \includegraphics[width=1.\textwidth]{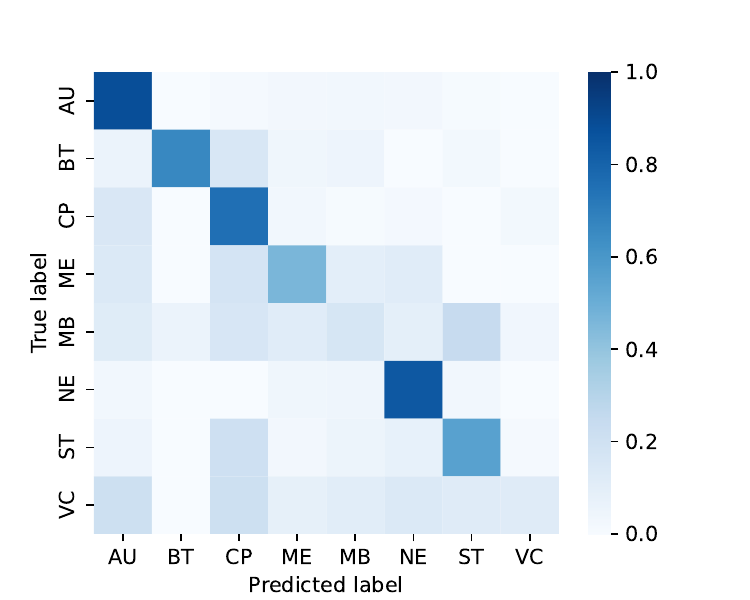}
        \subcaption{ZSL DV3X}\label{fig:cm-dv3x}
    \end{minipage}
    \begin{minipage}[t]{.32\textwidth}
        \centering
        \includegraphics[width=1.\textwidth]{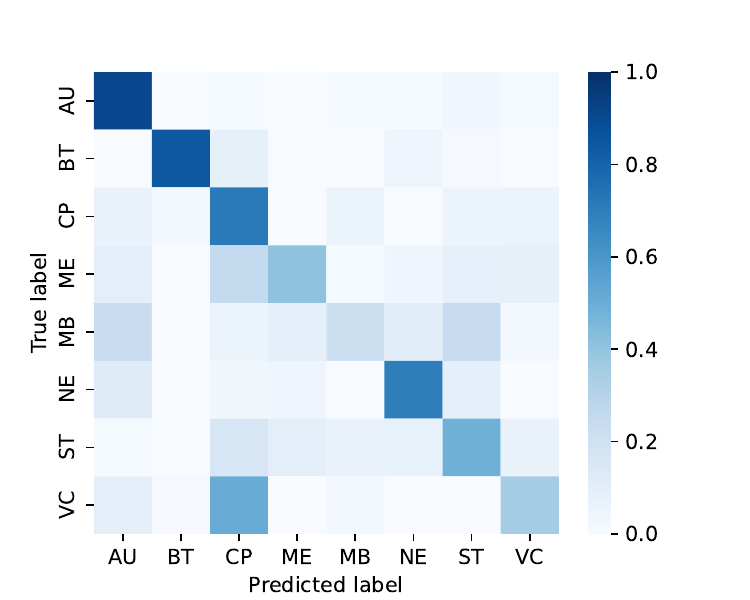}
        \subcaption{ZSL DV3T}\label{fig:cm-dv3t}
    \end{minipage}
    \begin{minipage}[t]{.32\textwidth}
        \centering
        \includegraphics[width=1.\textwidth]{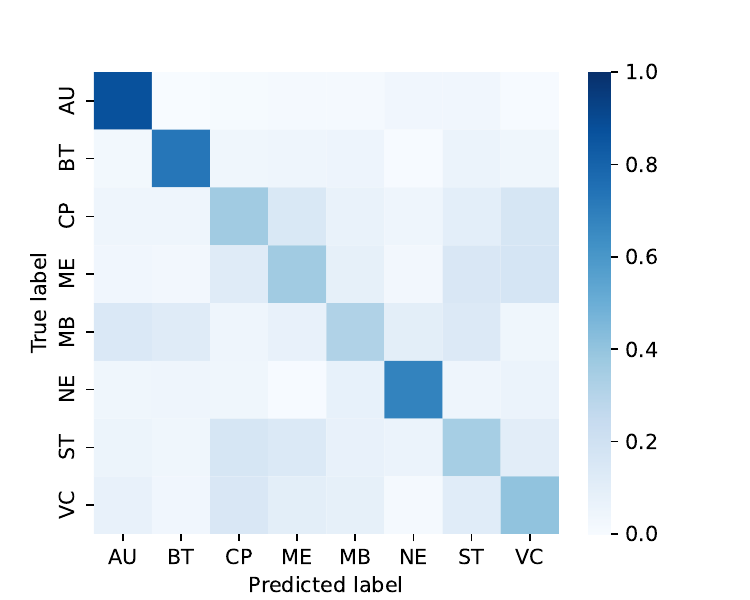}
        \subcaption{OSL 6MLM}\label{fig:cm-os-6mlm}
    \end{minipage}
    \hfill
    \vspace*{-3mm}
    \begin{minipage}[t]{.32\textwidth}
        \centering
        \includegraphics[width=1.\textwidth]{./figs/confusion_matrices/one_shot/6MLM.pdf}
        \subcaption{OSL 12MLM}\label{fig:cm-os-12mlm}
    \end{minipage}
    \begin{minipage}[t]{.32\textwidth}
        \centering
        \includegraphics[width=1.\textwidth]{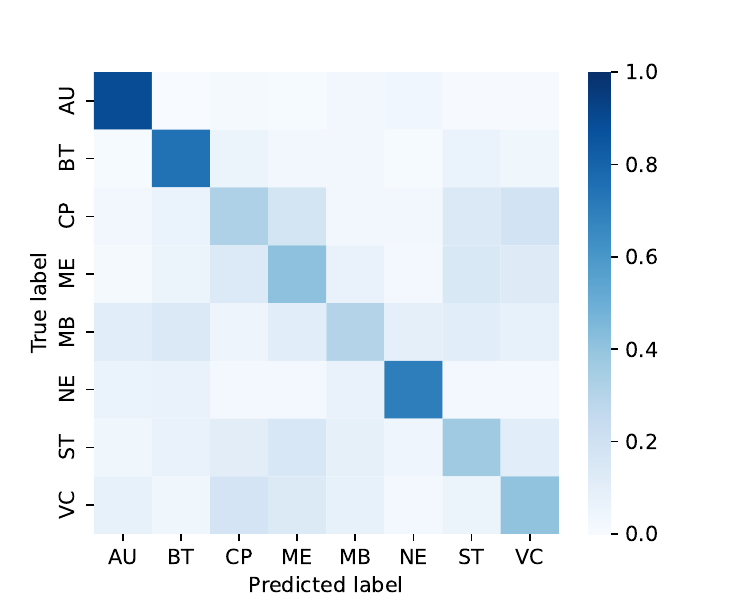}
        \subcaption{OSL MP}\label{fig:cm-os-mp}
    \end{minipage}
    \begin{minipage}[t]{.32\textwidth}
        \centering
        \includegraphics[width=1.\textwidth]{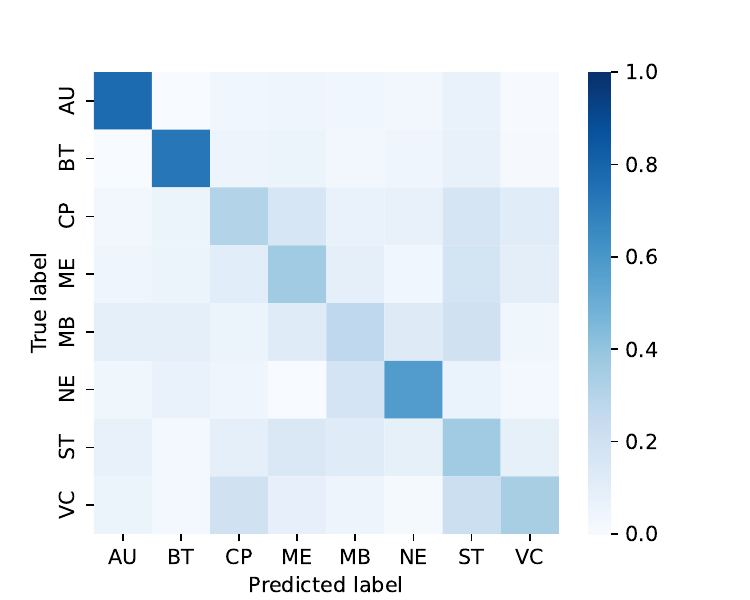}
        \subcaption{OSL MMP}\label{fig:cm-os-mmp}
    \end{minipage}
    \hfill
    \vspace*{-3mm}
    \begin{minipage}[t]{.32\textwidth}
        \centering
        \includegraphics[width=1.\textwidth]{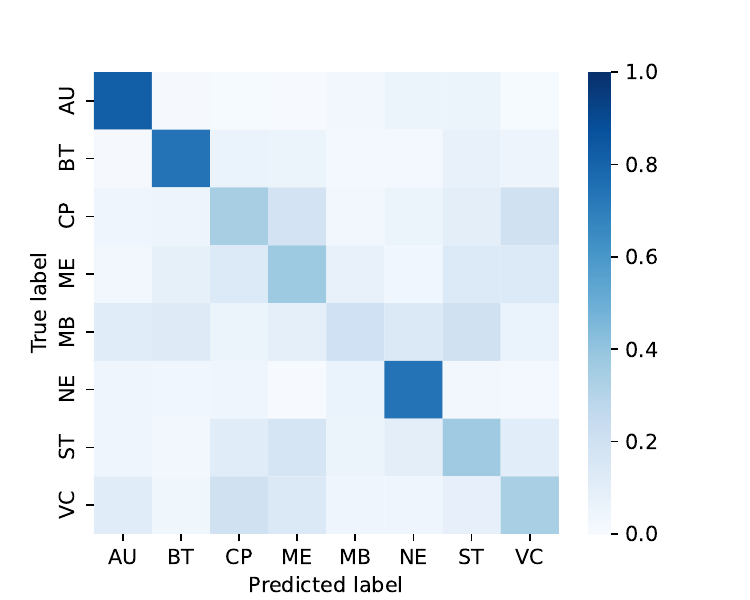}
        \subcaption{OSL MPQA}\label{fig:cm-os-mpqa}
    \end{minipage}
    \begin{minipage}[t]{.32\textwidth}
        \centering
        \includegraphics[width=1.\textwidth]{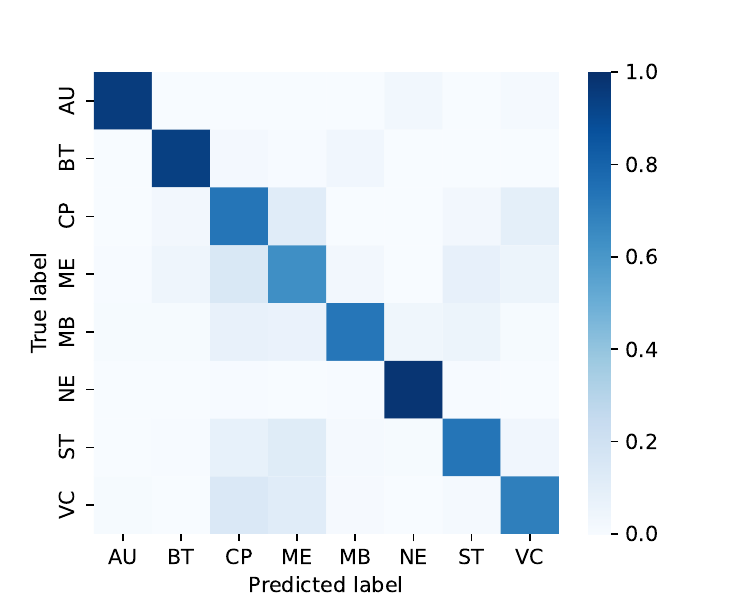}
        \subcaption{FSL 6MLM}\label{fig:cm-fs-6mlm}
    \end{minipage}
    \begin{minipage}[t]{.32\textwidth}
        \centering
        \includegraphics[width=1.\textwidth]{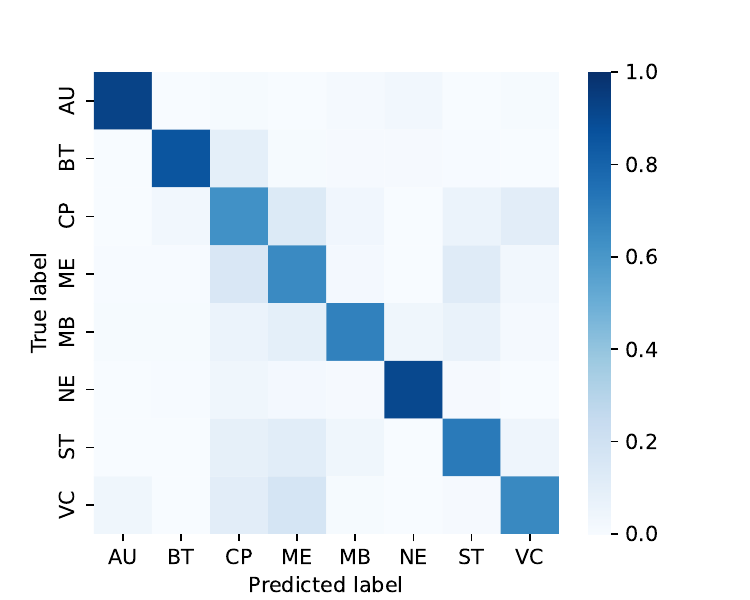}
        \subcaption{FSL 12MLM}\label{fig:cm-fs-12mlm}
    \end{minipage}
    \hfill
    \vspace*{-3mm}
    \hspace*{-5mm}
    \begin{minipage}[t]{.32\textwidth}
        \centering
        \includegraphics[width=1.\textwidth]{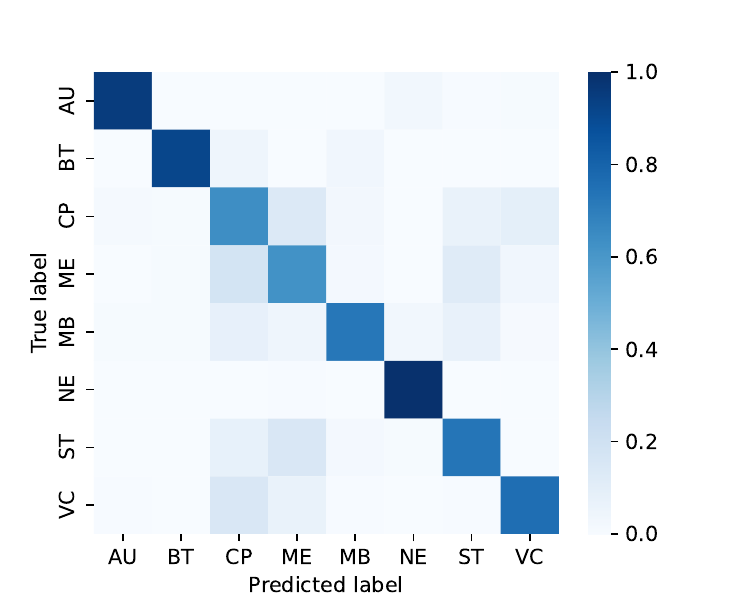}
        \subcaption{FSL MP}\label{fig:cm-fs-mp}
    \end{minipage}
    \begin{minipage}[t]{.32\textwidth}
        \centering
        \includegraphics[width=1.\textwidth]{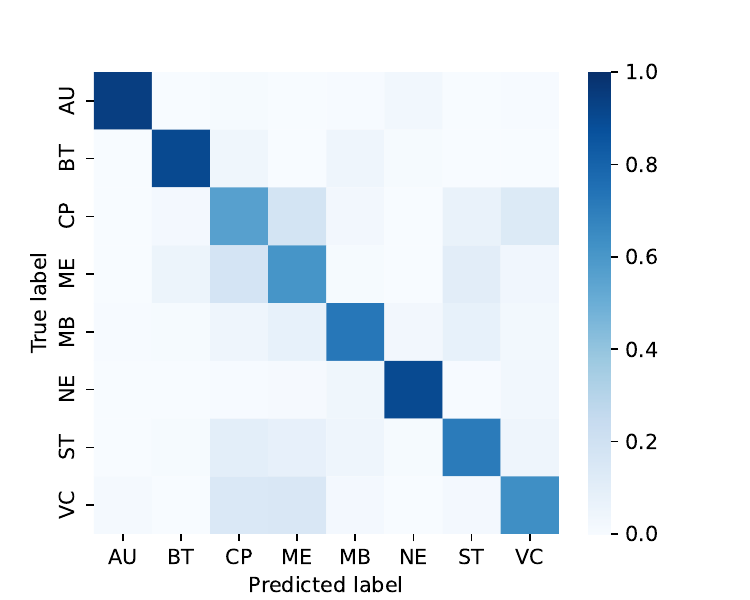}
        \subcaption{FSL MMP}\label{fig:cm-fs-mmp}
    \end{minipage}
    \begin{minipage}[t]{.32\textwidth}
        \centering
        \includegraphics[width=1.\textwidth]{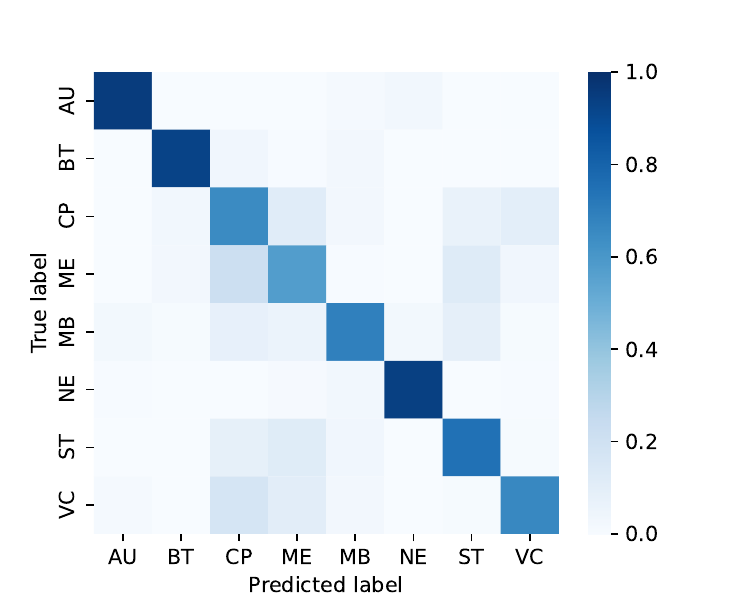}
        \subcaption{FSL MPQA}\label{fig:cm-fs-mpqa}
    \end{minipage}
    \vspace{-4mm}
\end{figure}

We now present the results of the NLP approaches applied for the classification task used to identify faulty computer components. The confusion matrices for each approach's predictions are shown in Table~\ref{fig:confusion-matrices}. The y-axis contains the true labels and the x-axis contains the predicted labels. Each cell is associated with a true label $true\_label$ and a predicted label $predicted\_label$ and contains the proportion number of samples of label $true\_label$ predicted as label $predicted\_label$ over the total number of samples of $true\_label$.

\begin{table}[t] \footnotesize
  \centering
  \caption{(a) ``User Report'' dataset sample. (b) Number of reports for each module in ``User Report'' dataset.}
  \begin{minipage}[t]{0.45\textwidth}
    \centering
    \resizebox{\textwidth}{!}{%
      \begin{tabular}{cc} \hline
         \textbf{Content} & \textbf{Label} \\ \hline
         I have a power issue. & Battery \\
         PC crashes when gaming. & CPU \\
         PC system does not boot up. & Memory \\ \hline
      \end{tabular}
    }
    \subcaption{}
    \label{tab:reports}
  \end{minipage}
  \hfill
  \begin{minipage}[t]{0.45\textwidth}
    \centering
    \resizebox{0.8\textwidth}{!}{%
      \begin{tabular}{ccc} \hline
             \textbf{Abbrev.} & \textbf{Module} & \textbf{\# Reports} \\ \hline
            VC & Video Card & 51 \\
            ST & Storage & 48 \\
            MB & Motherboard & 47 \\
            BT & Battery & 43 \\
            AU & Audio & 41 \\
            CP & CPU & 39 \\
            ME & Memory & 39 \\
            NE & Network & 33 \\ \hline
      \end{tabular}}
    \subcaption{}
    \label{tab:reports_values}
  \end{minipage}
  \vspace{-.5cm}
\end{table}

The more prominent the main diagonal, the more accurate the predictions made by the approaches. It is evident that the few-shot learning approaches yield better results compared to the others, as the models correctly predicted the classes of many samples. In zero-shot learning (ZSL) approaches the matrices are more dispersed. On the other hand, in one-shot learning (OSL) approaches, the main diagonal is lighter, for example, in approaches \ref{fig:cm-os-6mlm} and \ref{fig:cm-os-mmp}. 
There is an interesting observation: the zero-shot approaches seem to be better than the one-shot approaches. This behavior is explainable since the selected zero-shot models have more parameters than the other approaches, except MMP, as indicated in Table~\ref{tab:approaches-info}. This means the chosen zero-shot models are better than the one-shot models at generalizing the classes for this task.


In addition, there is a huge improvement when comparing one-shot to few-shot approaches. The few-shot approaches are expected to have better results due to the number of samples used to retrain the models. In our experiments, the few-shot methods use 23 samples of each class for the retraining phase, while the one-shot approaches use only one sample. The limited number of samples for one-shot classifiers was insufficient for effectively classifying the users' reports, potentially leading to a loss of predictive capability.

\begin{table}[b]
  \centering
  \caption{Results of accuracy and F1 score for each model.}
  \label{tab:metrics}
  \begin{minipage}[t]{0.32\textwidth}
    \centering
    \resizebox{\textwidth}{!}{%
      \begin{tabular}{ccc} \hline
         \textbf{Model} & \textbf{Accuracy} & \textbf{F1 Score} \\ \hline
         \textbf{BL} & \textbf{0.57} & \textbf{0.56} \\
         DV3 & 0.53 & 0.51 \\
         DV3M & 0.54 & 0.52 \\
         DV3T & 0.57 & 0.55 \\
         DV3X & 0.53 & 0.51 \\ \hline
      \end{tabular}
    }
    \subcaption{Zero-Shot}
    \label{tab:metrics_zero_shot}
  \end{minipage}
  \hfill
  \begin{minipage}[t]{0.32\textwidth}
    \centering
    \resizebox{\textwidth}{!}{%
      \begin{tabular}{ccc} \hline
         \textbf{Model} & \textbf{Accuracy} & \textbf{F1 Score} \\ \hline
         6MLM & 0.50 & 0.49 \\
         12MLM & 0.43 & 0.42 \\
         \textbf{MP} & \textbf{0.51} & \textbf{0.50}\\
         MMP & 0.45 & 0.44 \\
         MPQA & 0.47 & 0.44 \\ \hline
      \end{tabular}
    }
    \subcaption{One-shot}
    \label{tab:metrics_one_shot}
  \end{minipage}
  \hfill
  \begin{minipage}[t]{0.32\textwidth}
    \centering
    \resizebox{\textwidth}{!}{%
      \begin{tabular}{ccc} \hline
         \textbf{Model} & \textbf{Accuracy} & \textbf{F1 Score} \\ \hline
         \textbf{6MLM} & \textbf{0.79} & \textbf{0.79} \\
         12MLM & 0.75 & 0.75 \\
         MP & 0.78 & 0.79 \\
         MMP & 0.74 & 0.75 \\
         MPQA & 0.76 & 0.76 \\ \hline
      \end{tabular}
    }
    \subcaption{Few-shot}
    \label{tab:metrics_few_shot}
  \end{minipage}
\end{table}

Table~\ref{tab:metrics} shows the results considering the evaluation metrics. We highlighted the best approaches for our classification task for each paradigm. In zero-shot approaches, BL reached an accuracy and F1 score of $57\%$ and $56\%$, respectively, which is slightly better than the others. Also, the BL approach is the largest one between them with $407$ million parameters, more than twice the parameters of the other zero-shot models. Since BL is a larger model, it has more pre-trained knowledge encoded in its network to execute our task~\cite{brown2020language}. The one-shot approaches in Table~\ref{tab:metrics_one_shot} performed worse than the zero-shot, as we explained later. MP was the best one-shot approach with $51\%$ and $50\%$ of accuracy and F1 score, respectively. Curiously, the MiniLM with $6$ layers performed better than the MiniLM with $12$ layers. In this scenario, certain NLP tasks do not need the depth provided by a $12$-layer model, which can even introduce potential overfitting~\cite{goodfellow2016deep}. 





Among all the approaches evaluated, the few-shot 6MLM was demonstrated to be the best, with accuracy and F1 score of $79\%$ for both metrics. Although 6MLM has fewer parameters than the others, in cases where training data is limited or noisy, a simpler model might generalize better to unseen data since it is less likely to overfit the training data. Fewer parameters result in less capacity to memorize the training data, which helps the model to generalize better~\cite{goodfellow2016deep}. Besides that, the simpler model may reach its optimal performance quicker, allowing for more effective use of early stopping techniques to prevent overfitting~\cite{smith2018disciplined}.

\section{Conclusion and Future Work}
\label{sec:conclusion}

We conducted a study comparing different NLP models applied for a classification task to detect potential faulty computer components. Additionally, we designed a new dataset containing users' records reporting failures in their computers and the corresponding faulty module. The NLP models were retrained and tested using our dataset. We could observe that the number of samples used to retrain the models hardly affects the performance for this task. The few-shot models reached promissory results for preliminary research about the new context of faulty computer components.

For future work, we aim to improve and expand the application of NLP models to compose a framework for online assistance using large language models to detect faulty components with high accuracy. Additionally, we plan to extend this application to audio classification and integrate the models' prediction to smart speakers. Lastly, we aim to incorporate privacy features into NLP models.

\vspace{-.18cm}

\section*{ACKNOWLEDGMENT}
This research was partially funded by Lenovo, as part of its R\&D investment under Brazilian Informatics Law. 

\vspace{-.19cm}

\bibliographystyle{kdmile}
\bibliography{kdmileb}

\begin{received}
\end{received}

\end{document}